\pdfoutput=1
\documentclass[10pt,noshowpacs,aps,prb,a4paper,superscriptaddress,floatfix]{revtex4}
\usepackage{graphicx}
\usepackage{amsmath}
\usepackage{setspace}
\usepackage{amssymb}
\usepackage{CJK}
\usepackage{indentfirst}
\usepackage{threeparttable}

\par

\begin{document}
\title{Food Image Recognition by Using Convolutional Neural Networks (CNNs)}
\author{Yuzhen Lu}
\affiliation{Department of Biosystems and Agricultural Engineering, Michigan State University, East Lansing, MI 48824, USA; email: luyuzhen@msu.edu}

\begin{abstract}
\noindent Food image recognition is one of the promising applications of visual object recognition in computer vision. In this study, a small-scale dataset consisting of 5822 images of ten categories and a five-layer CNN was constructed to recognize these images. The bag-of-features (BoF) model coupled with support vector machine (SVM) was first evaluated for image classification, resulting in an overall accuracy of 56\%; while the CNN model performed much better with an overall accuracy of 74\%. Data augmentation techniques based on geometric transformation were applied to increase the size of training images, which achieved a significantly improved accuracy of more than 90\% while preventing the overfitting issue that occurred to the CNN based on raw training data. Further improvements can be expected by collecting more images and optimizing the network architecture and hyper-parameters.
\end{abstract}

\maketitle

\section{Introduction}
\noindent Due to the widespread use of low-cost imaging devices like smart-phone cameras, more and more applications are being developed in computer vision communities to facilitate automatic object recognition, among which recognition of food images has recently gained much attention \cite{ref1, ref2, ref3, ref4}. Nowadays, people, especially diabetes patients, are increasingly cautious about their diet for improved health care. Food image recognition provides a simple means to estimate the dietary caloric intake and evaluate eating habits of people, by using cameras to keep track of their food consumption.\par

\par\setlength\parindent{1em} In recent years, Convolutional neural networks (CNN) have enjoyed great popularity as a means for image classification/categorization since Krizhevsky et al \cite{ref5} won the ImageNet Large-Scale Visual Recognition Challenge (ILSVRC) 2012 competition. CNN, as a variant of the standard deep neural network (DNN), is characterized by a special network architecture consisting of alternating convolutional and pooling layers \cite{ref6}, to extract and combine local features from a two-dimensional input. Compared to conventional hand-crafted feature extraction based approaches, CNN is advantageous since it is able to learn optimal features from images adaptively. In implementing CNN for image classification, researchers have to collect such a large-scale dataset as the ILSVERC that contains more than one million images \cite{ref5, ref7}, for network training because of the need for learning a large number of parameters involved in the network, which, however, is not a trivial task. One simple way to deal with the situation is to apply the CNN model that has been pre-trained based on a large-scale image data \cite{ref1, ref8}, which is so-called transfer learning. Alternatively, one can algorithmically expand the existing training data, e.g., by performing affine transformations (e.g., translation, rotation, scaling, zooming, etc.) to the raw images \cite{ref9, ref10}.\par

\par\setlength\parindent{1em} This study was aimed to apply CNNs aided with data expansion techniques to a ten-class small-scale food image data. As comparison, a conventional a bag-of-feature (BoF) model combined with linear support vector machine (SVM) based approach was also employed for food image recognition. Experimental results demonstrated the superior performance of CNN and the effectiveness of data expansion techniques in training small-scale data.\par

\section{EXPERIMENTATION}
\subsection{Database}
\noindent A total of 5822 color images, representing ten-class food items, were collected from the ImageNet (http://image-net.org/). The number of images for each category is summarized in TABLE 1.

\begin{table}[htpb]
\footnotesize
\centering
\caption{Image number for each category of food items}
\begin{tabular}{c c c c c c c c c c c}
\hline
Category & Apple & Banana & Broccoli & Burger & Egg & Frenchfry & Hotdog & Pizza & Rice & Strawberry \\
\hline
Image number & 1050 & 310 & 327 & 519 & 626 & 296 & 639 & 1248 & 352 & 455\\
\hline
\end{tabular}
\end{table}

These images had a large variations in quality and size. Some of them have a neat, uniform background while some have cluttered background. Prior to image analysis, all the images were down-sampled to a fixed resolution of $128\times128$, and then randomly divided into two parts for training and test with a 4:1 ratio, resulting in 4654 training and 1168 test images.\par

\subsection{BoF}
\noindent The BoF model has been extensively used for image classification. In this method, an image is treated as a collection of orderless descriptors extracted from local patches, which are quantized into discrete visual words and represented by a compact histogram \cite{ref11}. Scale invariant feature transform (SIFT) descriptors that are not affected by perspective, scale, occlusion or illumination \cite{ref12}, are one of the most robust and popular feature descriptors. In this study, the BoF model with SIFT descriptors was used to extract features that were then fed to linear SVM for image classification, and this approach was implemented by means of the VLFeat library \cite{ref13}. \par

\subsection{CNN}
\noindent The architecture of the CNN used in this study is schematically illustrated in FIG. 1. The network has four layers of hidden neurons (three convolutional-pooling and one fully-connected), apart from a final layer of output neurons (the input is not considered as a layer). The input contains $128 \times128\times3$ neurons, representing the RGB values for a $128\times128\times3$ image. The first convolutional-pooling layer uses a local receptive field (also known as convolutional kernel) of size $7\times7$ with a stride length of 1 pixel to extract 32 feature maps, followed by a max pooling operation conducted in a $2\times2$ region; the second and third convolutional-pooling layers use $5\times5$ and $3\times3$ local receptive fields, resulting in 64 and 128 feature maps, respectively, and the other parameters remain unchanged. The fourth layer is a fully-connected layer with 128 rectified linear units (ReLU) neurons, and the output layer has 10 softmax neurons that correspond to the ten categories of food. The three convolutional-pooling layers also use ReLU activation functions. \par

\begin{figure}[htpb]
\begin{center}
\includegraphics[width=5.5in,height=1.6in]{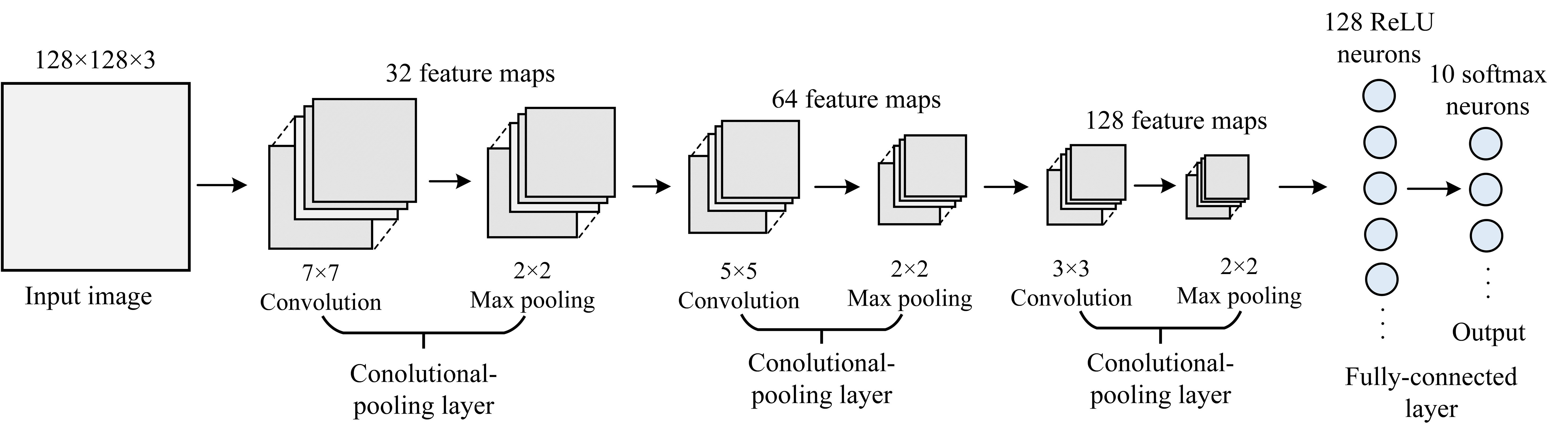}
\caption{Schematic of the architecture of a convolutional neural network model.}
\end{center}
\end{figure}

The network was trained with the stochastic gradient descent (SGD) algorithm with a cross-entropy cost function. The dropout that randomly eliminate a portion of neurons from the network was used to reduce possible overfitting. The dropout rates of 0.25 and 0.5 were set for the third convolutional-pooling layer and the fully-connected layer, respectively. Training a CNN model requires to select a set of hyper-parameters, among which the leaning rate $\eta$ is the the most critical one affecting the training performance. A fixed learning rate over the entire training process seems suboptimal, since it takes account of the dynamical training behavior. Here, a dynamically updated learning rate was used, which was an exponential function of cost $\eta=\eta_0\times\exp(C)$ where $\eta_0$ is set to 0.00l through trials and errors and C is the training loss. Such learning rate updating schedule is directly related to the training performance. At the initial stage, the training loss was large, resulting in a large learning rate to speed up the training process; gradually, the learning rate decreased with the loss, which helped avoid overshooting the best result.
Given the limited images available, affine transformations including rotation, translation and scaling were conducted to expand the training data. Upper bounds were set up in performing these transformations, within which each image was randomly subjected to those transformations, which substantially increased the training data size. FIG. 2 shows an example of a raw image and expanded images. \par

\begin{figure}[htpb]
\begin{center}
\includegraphics[width=5.2in,height=2.7in]{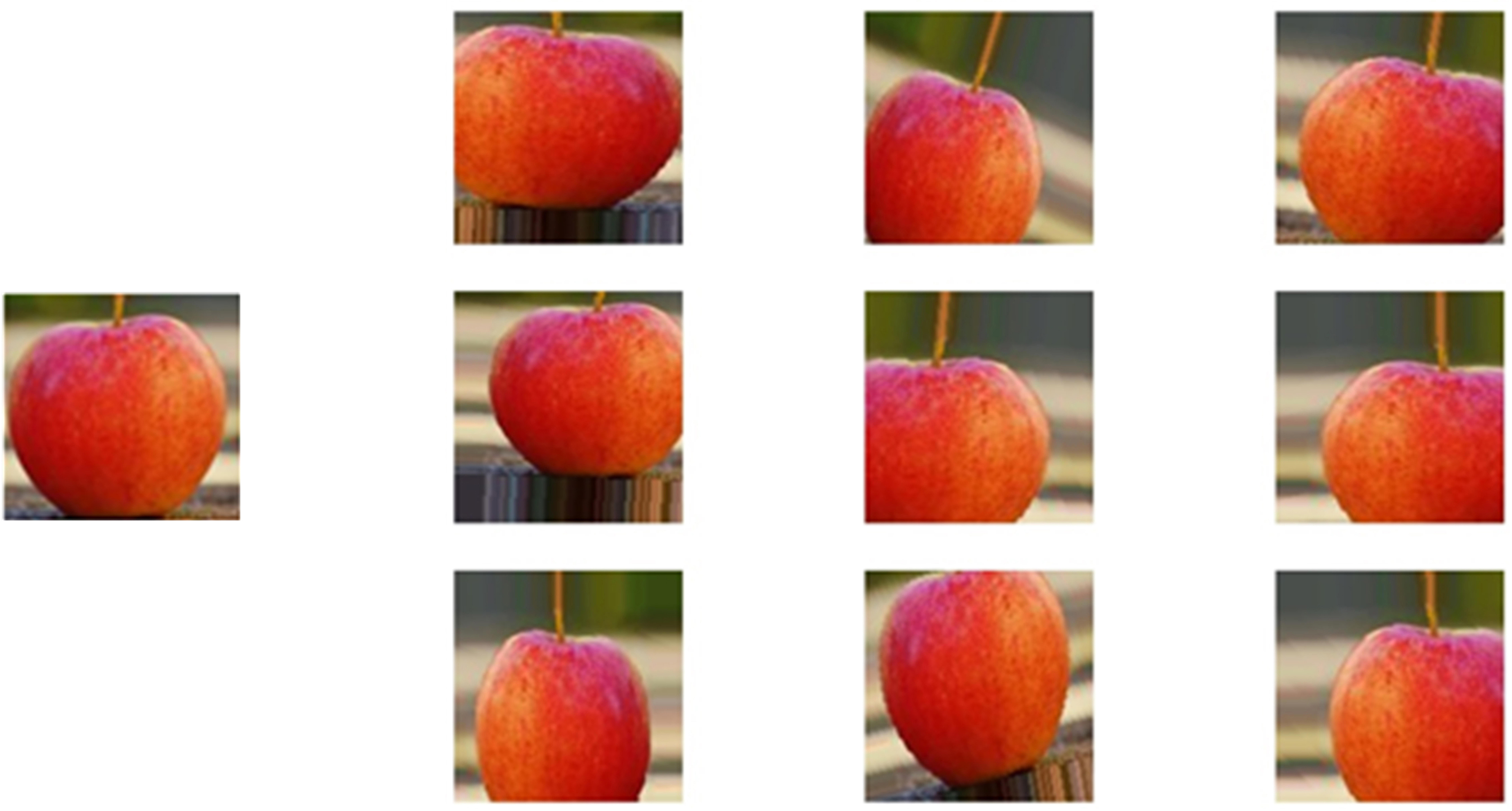}
\caption{The raw fruit image (right) and expanded images (left).}
\end{center}
\end{figure}

The CNN model was trained by using deep learning packages of Keras (https://github.com/keras-team) and Theano (https://github.com/Theano)(the implementation codes are available at: https://github.com/jingweimo/food-image-classification-).\par

\section{RESULTS AND DISCUSSION}
\noindent The BoF combined with SVM resulted in an overall accuracies (the ratio of the number correctly recognized images to the number of total images) of 68\% and 56\% for training and test images, respectively. TABLE 2 presents the confusion matrix on the test images and recognition rates (i.e., the average of true positive and true negative rates) for all the categories.\par

\begin{center}
\begin{threeparttable}
\centering
\caption{Confusion matrix and recognition rates by BoF on the test images}
\begin{tabular}{c c c c c c c c c c c}
\hline
Category & Apple & Banana & Broccoli & Burger & Egg & Frenchfry & Hotdog & Pizza & Rice & Strawberry \\
\hline
Apple & \bf{178} & 1 & 0 & 4 & 8 & 0 & 2 & 6 & 5 & 7\\
Banana & 2 & \bf{43} & 1 & 2 & 4 & 4 & 5 & 2 & 1 & 1\\
Broccoli & 1 & 0 & \bf{28} & 2 & 0 & 3 & 2 & 24 & 1 & 4\\
Burger & 5 & 0 & 2 & \bf{72} & 2 & 2 & 7 & 12 & 1 & 1\\
Egg & 20 & 1 & 2 & 7 & \bf{75} & 1 & 6 & 6 & 6 & 1\\
Frenchfry & 1 & 4 & 3 & 4 & 1 & \bf{21} & 6 & 16 & 1 & 0\\
Hotdog & 5 & 8 & 4 & 10 & 4 & 9 & \bf{76} & 10 & 2 & 1\\
Pizza & 5 & 0 & 4 & 5 & 1 & 1 & 7 & \bf{221} & 3 & 4\\
Rice & 6 & 1 & 1 & 2 & 4 & 1 & 1 & 18 & \bf{35} & 1\\
Strawberry & 11 & 1 & 2 & 1 & 1 & 0 & 1 & 29 & 1 & \bf{45}\\
R.R. & 0.89 & 0.82 & 0.71 & 0.83 & 0.79 & 0.67 & 0.78 & 0.87 & 0.74 & 0.73\\
\hline
\end{tabular}
\begin{tablenotes}[flushleft]\footnotesize
\small
\item R.R. denotes the recognition rate (i.e, the average of the true positive and true negative rates).
\end{tablenotes}
\end{threeparttable}
\end{center}

\bigskip
Apple and pizza gave the two highest recognition rates, which were mainly because the two categories had a large number of training images; while french-fry and broccoli was the two hardest categories, the majority of which were misclassified into another distinct class. The BoF based approach did not performed well as expected, which was possibly because that the extraction of SIFT descriptors was conducted only to grayscale images, and also the parameter optimization was not fully done in descriptor quantization.
The CNN was first implemented without using data expansion techniques. FIG. 3 show the accuracy and loss curves during training with the maximum training epoch set to 100. A large gaps between training and test occurred after 10 epochs, indicating the presence of overfitting. The highest accuracy on the test images was found to be 74\%, corresponding to a training accuracy 95\%. These results were much better than those obtained by the BoF approach, which confirmed the superiority of CNN.\par

\begin{figure}[htpb]
\begin{center}
\includegraphics[width=5.6in,height=2in]{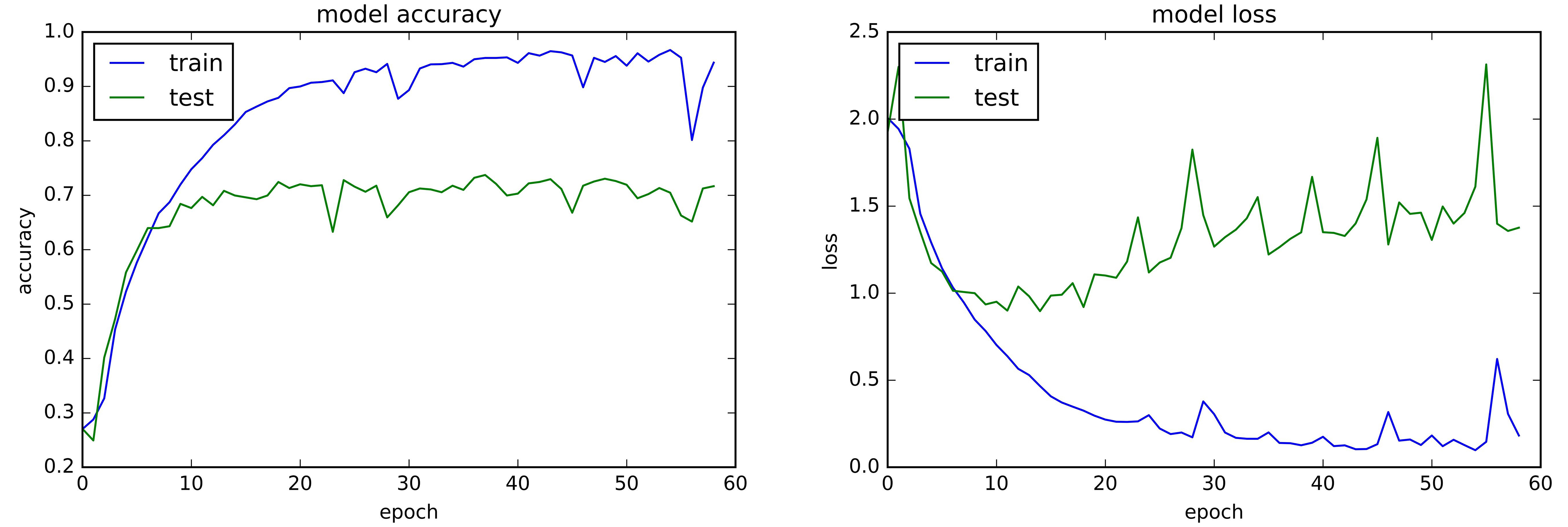}
\caption{Accuracy (left) and loss (right) curves for convolutional neural network models without data augmentation. The training process was early stopped because of no improvement over a preset period of time (20 epchos in this study).}
\end{center}
\end{figure}

Then, the CNN was trained with the expanded image data. As illustrated in FIG. 4, two significant improvements were achieve through almost trivial data transformations. First, the test accuracy was greatly elevated to a level up to 87\% within 100 epochs, and second, the overfitting issue observed above was completely eliminated. The data expansion substantially increases the effective size of the training data, thus helping improve the training performance and making the model generalize well.\par

\begin{figure}[htpb]
\begin{center}
\includegraphics[width=5.6in,height=2in]{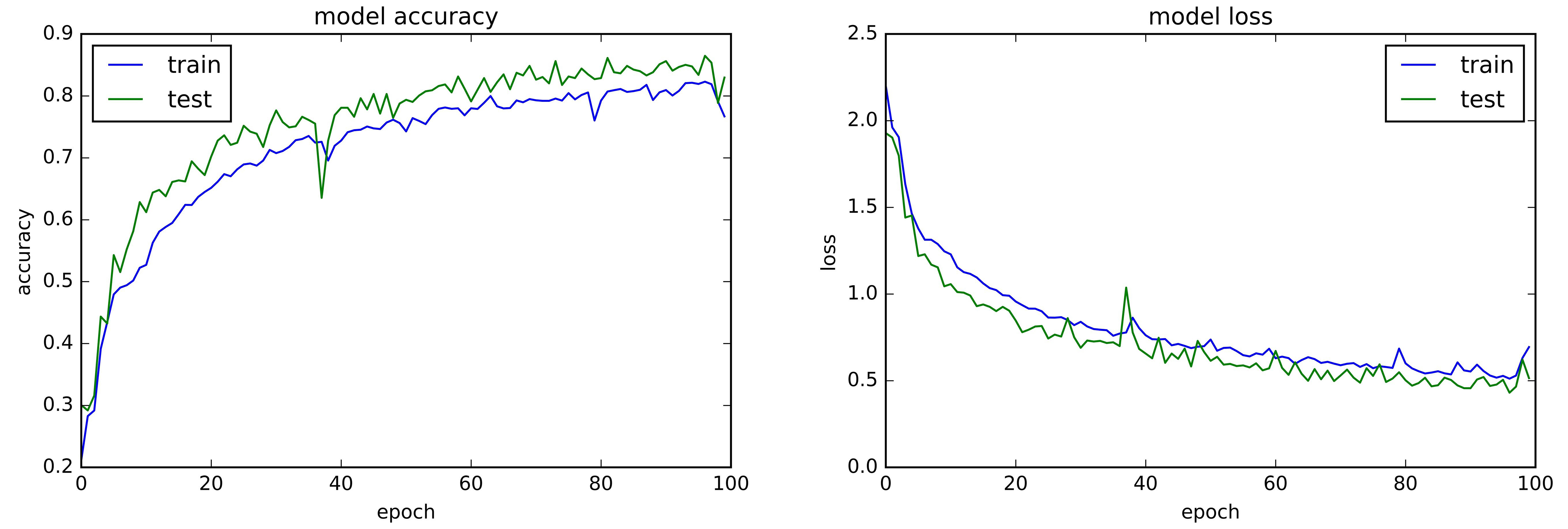}
\caption{Accuracy (left) and loss (right) curves for convolutional neural network models based on data augmentation.}
\end{center}
\end{figure}

According to the accuracy and loss curves, it seemed that the CNN model could be further improved by increasing the training epochs. Hence, another three training schemes with 200, 400 and 600 epochs respectively were implemented. FIG. 5 shows their accuracy and loss curves. The test accuracy did increase further, but not by a large margin. After 200 epochs the training accuracy started to exceed the test accuracy and the gaps between training and test tended to enlarge. Training the CNN with 400 epochs resulted in the highest test accuracy of more than 90\%; while training 600 epochs only increased the training accuracy, but not the test accuracy which seemed to level off around 90\%. This suggested that overfitting was possible to escalate, and one may not further increasing the test accuracy by simply increasing the training epochs. While data augmentation is an effective method for improving the CNN performance, it does not overshadow the importance of manually collecting more training images.\par

\begin{figure}[htpb]
\begin{center}
\includegraphics[width=5.6in,height=6in]{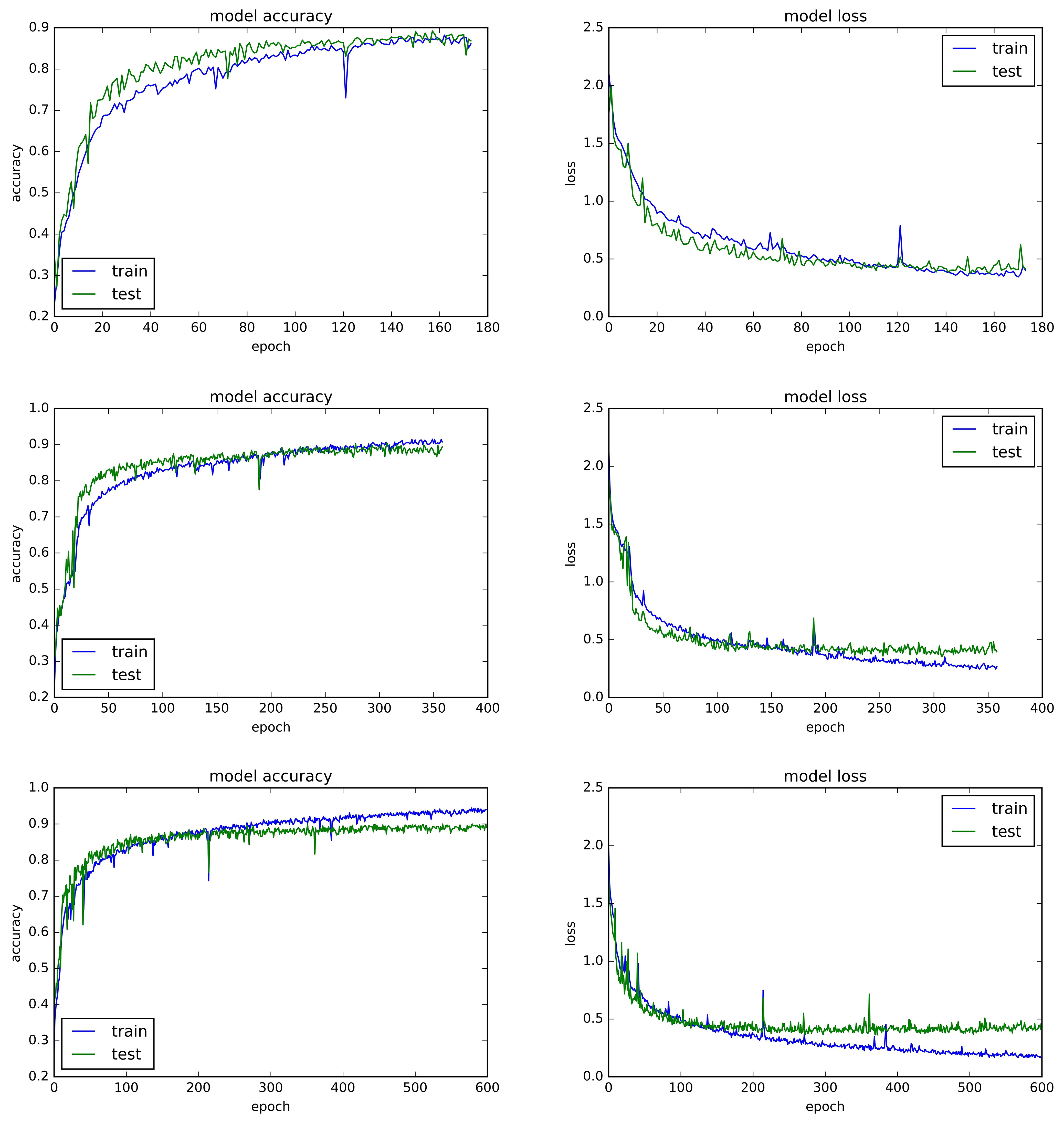}
\caption{Accuracy (left) and loss (right) curves for convolutional neural network models with training epochs set to 200, 400 and 600 (from top to bottom), where early stopping could occur because of no accuracy improvement on test images.}
\end{center}
\end{figure}

TABLE 3 presents the confusion matrix by the best CNN model for the test images and the corresponding recognition rates for each food category. Compared with Table 2, the CNN model resulted in the overall accuracy improved from 56\% to 90\%, and the average recognition rate from 0.78\% to 94\%.\par

\begin{center}
\begin{threeparttable}
\centering
\caption{Confusion matrix and recognition rates by BoF on the test images}
\begin{tabular}{c c c c c c c c c c c}
\hline
Category & Apple & Banana & Broccoli & Burger & Egg & Frenchfry & Hotdog & Pizza & Rice & Strawberry \\
\hline
Apple & \bf{193} & 6 & 1 & 0 & 1 & 0 & 2 & 1 & 0 & 6\\
Banana & 4 & \bf{49} & 0 & 0 & 4 & 2 & 3 & 0 & 0 & 0\\
Broccoli & 0 & 0 & \bf{64} & 0 & 0 & 0 & 1 & 1 & 0 & 0\\
Burger & 1 & 0 & 0 & \bf{87} & 0 & 1 & 9 & 6 & 0 & 0\\
Egg & 3 & 1 & 0 & 2 & \bf{110} & 2 & 5 & 1 & 2 & 0\\
Frenchfry & 0 & 2 & 0 & 0 & 0 & \bf{53} & 1 & 4 & 0 & 0\\
Hotdog & 1 & 2 & 0 & 5 & 0 & 3 & \bf{109} & 8 & 0 & 0\\
Pizza & 0 & 0 & 0 & 3 & 0 & 1 & 6 & \bf{239} & 0 & 1\\
Rice & 0 & 0 & 0 & 1 & 0 & 0 & 1 & 5 & \bf{64} & 0\\
Strawberry & 3 & 0 & 0 & 0 & 0 & 0 & 0 & 0 & 0 & \bf{88}\\
R.R. & 0.95 & 0.89 & 0.98 & 0.91 & 0.93 & 0.94 & 0.91 & 0.96 & 0.95 & 0.98\\
\hline
\end{tabular}
\begin{tablenotes}[flushleft]\footnotesize
\small
\item R.R. denotes the recognition rate (i.e., the average of the true positive and true negative rates).
\end{tablenotes}
\end{threeparttable}
\end{center}

\section{Conclusions}
\noindent This study reported on the applications of CNNs to a ten-class small-scale food image data. A five-layer CNN model was constructed achieved the best test accuracy of 74\%, which was better than the accuracy of 56\% achieved by the BoF approach. The CNN model however suffered from noticeable overfitting due to limited training data. This issue was addressed by expanding the training data through various affine transformations, which also considerably increased the overall test accuracy to more than 90\%. Training the CNNs with different epochs showed limited room for improving test accuracy. Further improvements can be anticipated by collecting more training data and or by optimizing the architecture and hyper-parameters of the network, rather than by based on increasing the training epochs in the current framework, which, otherwise, could be liable to overtraining.\par

\end{document}